\RequirePackage{fix-cm}
\documentclass[smallextended,natbib]{svjour3}       
\smartqed  
\usepackage{parskip}
\usepackage{graphicx}

\usepackage{caption} 
\usepackage{subcaption}
\usepackage{amsmath}
\usepackage{amsfonts}

\usepackage{enumitem}
\usepackage{url}
\begin{document}

\title{Definitions of intent suitable for algorithms\thanks{Supported by EPSRC}
}


\author{Hal Ashton       
}


\institute{H. Ashton \at
              University College London \\
              \email{ucabha5@ucl.ac.uk}
}

\date{Received: date / Accepted: date}

\maketitle

\begin{abstract}
Intent modifies an actor's culpability of many types wrongdoing. Autonomous Algorithmic Agents have the capability of causing harm, and whilst their current lack of legal personhood precludes them from committing crimes, it is useful for a number of parties to understand under what type of intentional mode an algorithm might transgress. From the perspective of the creator or owner they would like ensure that their algorithms never intend to cause harm by doing things that would otherwise be labelled criminal if committed by a legal person.  Prosecutors might have an interest in understanding whether the actions of an algorithm were internally intended according to a transparent definition of the concept. The presence or absence of intention in the algorithmic agent might inform the court as to the complicity of its owner. This article introduces definitions for direct, oblique (or indirect) and ulterior intent which can be used to test for intent in an algorithmic actor.
\end{abstract}

\section{Introduction}\label{intro}
The establishment of criminal intent, the volitional component of mens rea in the someone accused of committing a crime, is a necessary task in proving that a crime was committed. An autonomous algorithm with agency can perform actions which would considered criminal (actus reus) if a human with sufficient mens rea performed them. For brevity in this article we will follow \cite{Abbott2020PunishingFiction} and henceforth refer to this state of affairs as an \textbf{AI-crime}. It is entirely possible that an autonomous algorithm can commit a AI-crime, with out having being instructed to by its programmers or owners, it instead learned to behave that way through some machine learning technique. Abbott and Sarch term this a hard AI crime, distinguishing it from where an AI is designed and used as a tool to commit crime on behalf of its owners or creators. 
Hard AI-crimes admit the possibility of a responsibility gap appearing, harms might be committed but with no-one criminally responsible for them. We will refer to this class of self-learning, autonomous algorithms as A-bots for the remainder of this article.


\begin{figure}
    \centering
    \includegraphics[width=0.9\textwidth]{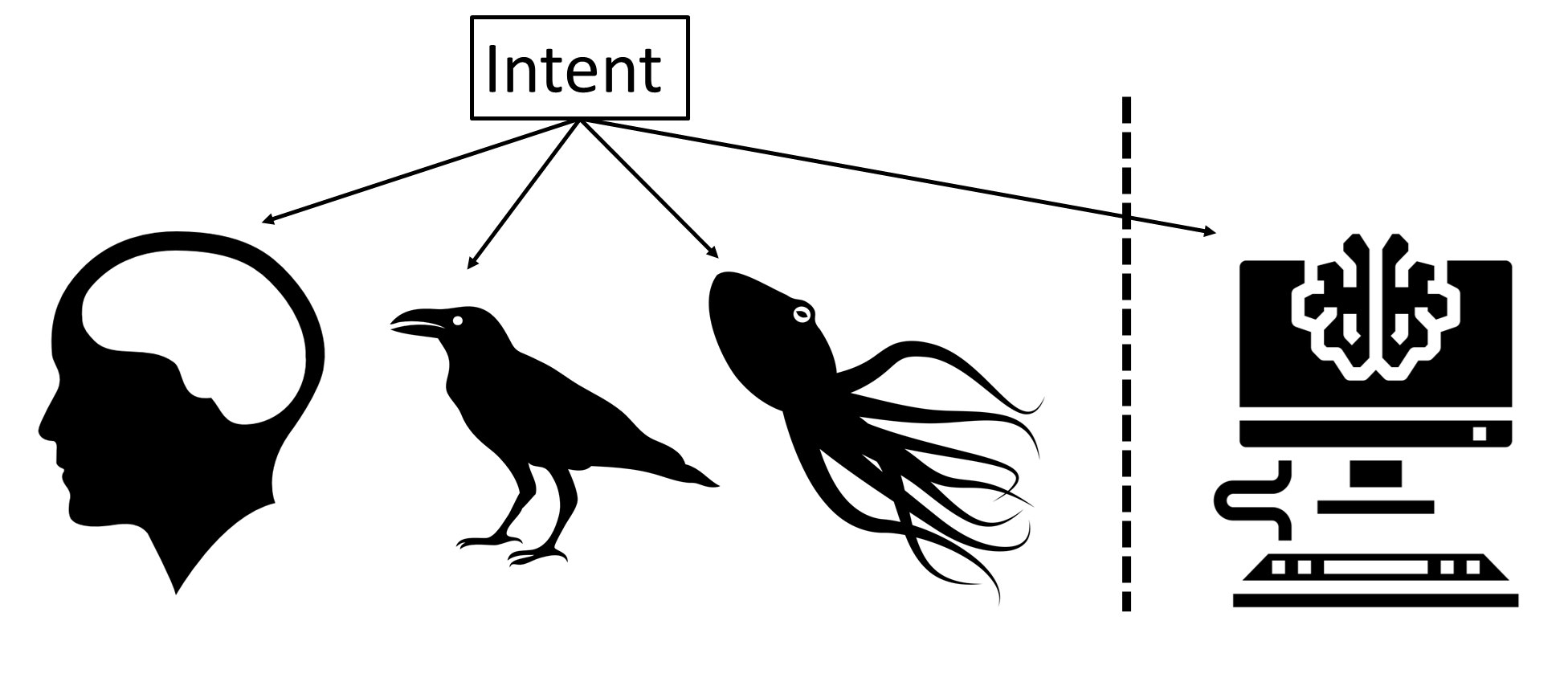}
    \caption{This paper proceeds under the assumption that intent is a definable concept that does not require a human brain to exist, that it arguably exists in other biological entities with demonstrable intelligence and can plausibly exist in an artificial intelligence. \small\textit{Images: Octopus - James Keuning, AI - Komkrit Noenpoempisut, The Noun project}
}
    \label{fig:my_label}
\end{figure}

If the owner or creator of an algorithm intended their algorithm to commit a AI-crime, and it did subsequently do so, then the owner or creator (hence the principal) would be guilty of the crime in the same way as anyone using a tool to commit a crime is. The doctrine of innocent agency goes further, and prevents the Principal from using other people as tools\footnote{Assuming that the person used as a tool is not aware that they are committing a crime.} to commit a crime on their behalf. Conversely if the Principal did not obviously intend their algorithm to commit a AI-crime, their culpability becomes less clear. Arguments could be made about the foreseeability of a AI-crime being committed by the algorithm on behalf of the Principal. That might be sufficient to establish lower levels of intent like negligence in the Principal, but it might not be when judging whether they had higher levels of intent which are required for the more serious crimes of specific intent. If a general definition of intent existed for algorithms, then it would be harder for a principal to argue that they did not know that an algorithm intended to commit a AI-crime. Wilful blindness as to a fact has been established, under certain circumstances, to be equivalent to knowledge of a fact ('The Ostrich instruction' - \cite{Robbins1990TheRea}). An algorithmic definition of intent might not allow one to conclude that intent in the AI equals intent in the Principal, but at the same time it might be useful evidence as to the intentional state of the Principal as to their algorithm. The idea of criminal responsibility being somewhat transferable from Agent back to Principal is not an alien idea to criminal law and is a concept known as secondary liability. A Principal that knows (or should know) that their agent will likely commit a crime can be considered an abettor or accessory to that crime. However at present, abetting or encouraging is parasitic on the Agent's crime \citep{Kaiserman2021AgainstLiability}, that is to say without the Agent committing a crime, there can be secondary criminal liability placed on the Principal. Since AI-crimes are not crimes, then the Principal would not be liable under this mechanism.  

From the perspective of the algorithm creator, a practical definition of intent, compatible with some control mechanism, is useful under the assumption that they would like their algorithm not to intend to commit AI crimes. Aside from all criminal law requiring an intentional state, accompanied with a proscribed behaviour, this article will show other areas of law such as Contract law \cite{Ayres2005InsecureIntent}, Tort \citep{Cane2019MensLaw} and Regulatory law \citep{YavarBathaee2018TheCausation} have occasion to establish intent. Sometimes, the behaviour might not even be obviously proscribed in isolation of the intentional state.  

The approach of this article is atypical in computer science literature in that the definitions of intent that it will present are informed by the body of law that exists. Other approaches might be to use psychological evidence or ethical theory. However, we believe a theory built on either of those alternatives has a lower chance of protecting programmers from legal action than one which is founded on legal precedent. We are also wary about the normative effect on law that such approaches have for there can be a democratic deficiency in them as \cite{Hildebrandt2019Closure:Law} points out. From the perspective of the development of the law, AI poses genuinely novel challenges. To quote Lord Mance \footnote{Quoine Pte Ltd v B2C2 Ltd [2020] SGCA(I) 02 at 193}:
\begin{quote}
    ...the  law must  be adapted to the new algorithmic programmes and artificial intelligence, in a way which gives rise to the results that reason and justice would lead one to expect
\end{quote}

Hard AI-crimes are likely to appear before courts before sufficient statutory laws are made, so existing precedent will go a long way to informing how intent is treated in A-bots. The article is roughly divided in two parts and will proceed as follows. Firstly in Section \ref{sec:legal} we will consider various different types of intent that exist in criminal law and their definitions such as they are. We will also venture briefly outside criminal law to see where intent is required. Armed with that knowledge, Section \ref{sec:algo_defn} will discuss what is required in an A-bot for intent to exist, some desiderata of intent definitions and finally definitions of Direct, Oblique and Ulterior intent. This is followed by a short discussion and a review of alternative attempts to formally define intent.

\section{Definitions of Intent from legal literature}\label{sec:legal}
\begin{figure}[!ht]
    \centering
    \includegraphics[width=0.7\textwidth]{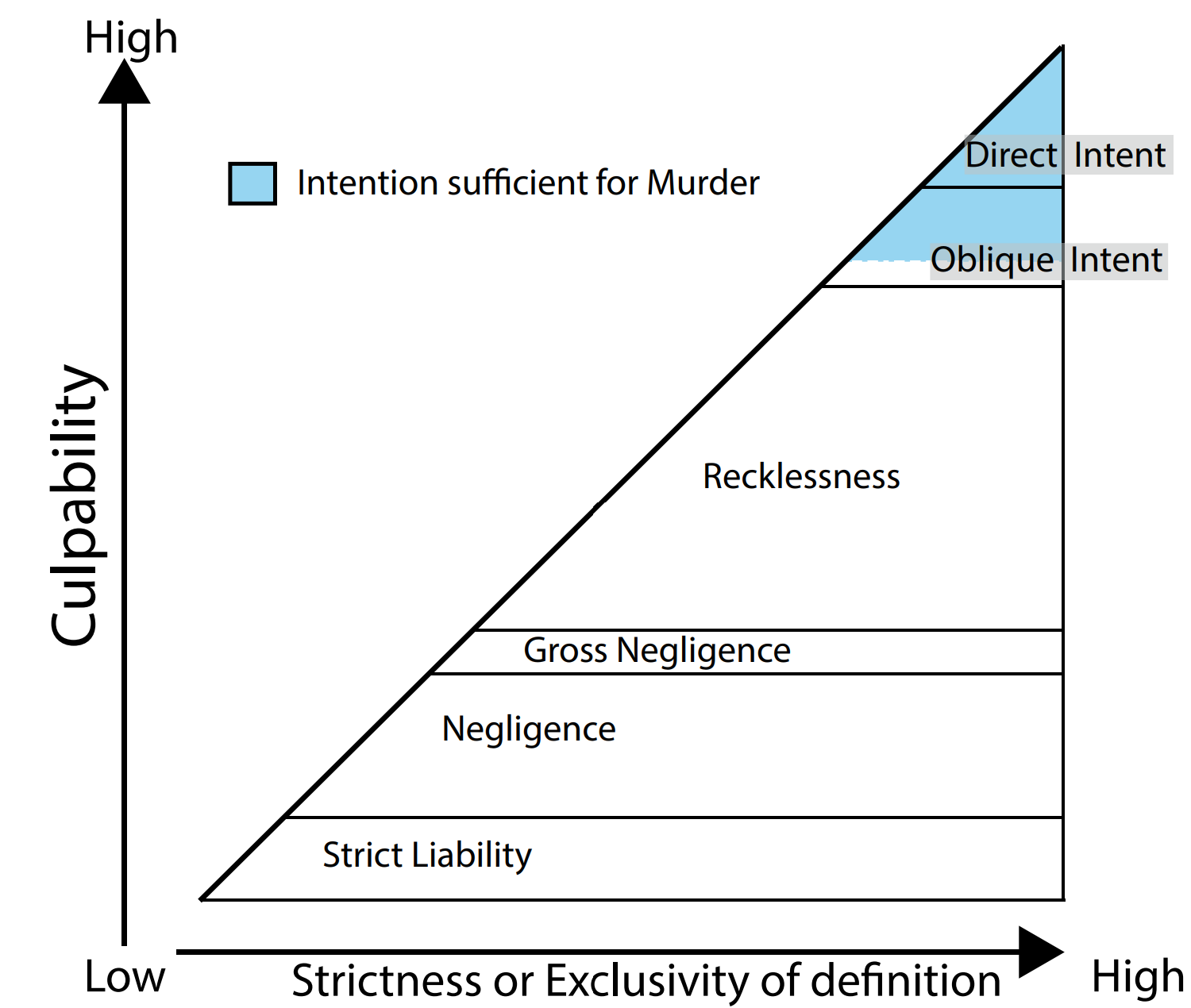}
    \caption{The Intent hierarchy and its relationship with culpability. Adapted from \cite{Loveless2010MensNegligence}}
    \label{fig:triangle}
\end{figure}

Intent within a criminal law context is generally arranged in a hierarchy; different modes of intent are required to meet increasingly strict definitions. Specific crimes are typically defined with a threshold level of criminal intent; the minimum level of intent that the accused must have in order to have committed the mental element of the crime. One reason for this is that increased levels of intent behind a criminal act correspond to increased levels of culpability and sanction. The clearest example of this is with the offence of murder; if the act of killing someone is done with direct intent then it is murder, if death is a result of lower intentional mode such as recklessness, then it would be manslaughter \footnote{This is a simplification, in the UK there are further distinctions between voluntary and involuntary manslaughter  \citep{Service2019Homicide:Manslaughter} and as we will discuss oblique intent \emph{can} be sufficient for murder. }. This is shown in figure \ref{fig:triangle}; different modes are arranged hierarchically, the x-axis representing the exclusivity of strictness of their definition, and the y-axis representing the culpability that society assigns to acts committed with those levels of intent. An action committed with a higher mode of intent such as Direct Intent, is also therefore committed with Recklessness\footnote{See for example cl19 of \cite{Commission1989ABill}}. Another justification for establishing the intent behind an action is to distinguish between those outcomes which were accidental and those which were not. Sometimes only proof of harm is required irrespective of outcome; this is called strict liability and forms the lowest level of the hierarchy. It should be noted that there is no universal language for intent across nations and justice systems, so concepts Negligence or Recklessness might mean different things in different places or may have analogous modes with other names.   

The original intention of this paper was to exclusively concentrate on the highest modes of intent - Direct and Oblique - because there are many commonalities across the world and between disciplines, as to what these concepts constitute. However we have included discussion of Recklessness and Negligence, because we have found them useful to discuss what the higher levels of intent are and are not. Where intent is mentioned, the reader should assume direct and oblique intent are the subject of the discussion. It will be made clear when the lower modes of recklessness and negligence are also being discussed. 

\subsection{Intent in common law}
A barrier to creating a legally rigorous algorithmic definition of intent is that courts in the UK have consistently not wanted to elaborate to juries what intent actually constitutes. As Lord Bridge stated \emph{"The judge should avoid any elaboration or paraphrase of what is meant by intent and leave it to the jury's good sense to decide whether the accused acted with necessary intent"}\footnote{R v Moloney (1985) 1 All ER 1025.}. A potential reason behind this is the confounding existence of oblique (sometimes called indirect intent), which whilst occupying a lower level to direct intent in the hierarchy of Figure \ref{fig:triangle}, has been established in a number of boundary cases such as R v Nedrick\footnote{R v Nedrick [1986] 1 WLR 1025.} and R v Woollin\footnote{\label{woollin}R v Woollin [1999] 1 A.C. 82.} to be sufficient, in certain cases, to be sufficient mens rea for the most serious crime of murder. We will discuss Oblique intent after tackling direct intent.

\subsubsection{Direct Intent}

Whilst a definition of direct intent has not been forthcoming within courts in the UK, examples do necessarily exist within textbooks and other legal discourse\footnote{Else how would anyone know what needs to be proven in a criminal court.}. \cite{Parsons2000IntentionFind} defines direct intent as the case where \emph{"the defendant wants something to happen as a result of their conduct"}. A draft bill published by the UK Home Office \cite{Commission2015AppendixBill} defines direct intent as the situation when \emph{A person acts intentionally with respect to a result if...it his purpose to cause it}. Using this document as a consultation template, the Law commission \cite{TheLawCommission2015ReformReport} also suggested an alternative formulation of direct intent as follows:
\begin{quote}The jury should be directed that they may find D intended a result if they are sure that D realised that result was certain (barring an extraordinary intervention) if D did what he or she was set upon doing.
\end{quote} A previous formulation is to be found in a draft criminal code \\ \cite{Commission1989ABill}, which states that 

\begin{quote}
A person acts intentionally with respect to i) a circumstance when he hopes or knows that it exists or will exist; ii) a result when he acts either in order to bring it about or being aware that it will occur in the ordinary course of events. 
\end{quote} 

It should be noted that the Law Commission's 2015 consultation concludes that no definition is needed, at least in the context of the offences against the person bill reform. 

As \cite{Coffey2009CodifyingLaw} summarises, the ingredients of direct intent generally seem to involve a decision to act and an outcome which is the aim, objective or purpose of that act. Whether that outcome or result is desirable from the point of view of the accused seems to depend on the narrowness of the definition of desire. On the subject of direct intent,  James LJ in R v Mohan [1976] 1 QB at 11 says it is: 
\begin{quote}
...a decision to bring about insofar as it lies within the accused's power, the commission of the offence which it is alleged the accused attempted to commit, no matter whether the accused desired that consequence of his act or not.
\end{quote}

In the USA, a definition of direct intent is more forthcoming in the form of the Model Penal Code (MPC). This has been adapted to various degrees by many states, though Federal prosecuted crimes have no analogous written definitions. What we have termed direct intent corresponds to the MPC's definition of purpose, the highest of the four levels of intent that they define \citep{TheAmericanLawInsitute2017GeneralCulpability}:

\begin{quote}
    A person acts purposely with respect to a material element of an offense when... if the element involves the nature of his conduct or a result thereof, it is his conscious object to engage in conduct of that nature or to cause such a result
\end{quote}

Generally we can conclude that directly intended things do not need to be desirable but they should be aimed for. The example of a dentist is often given to illustrate this point \citep{Williams1987ObliqueIntention}. A painful tooth extraction may result, which is certainly not desirable for most, but the object of the visit is to obviate future tooth ache\footnote{The intentional state of the pain that necessarily ensues is discussed in the next sub-section.} 

Related, and sometimes confused with oblique intent, is the intentional status of intermediate results which are caused through the actions of the agent, and are necessary to achieve some other aimed for result. These intermediate results, which \cite{Simester2019MensRea} term \emph{Means to an end} results, are directly intended, this being established in Smith [1960] 2 QB 423 (CA) where it was found that a defendant who bribed a Mayor in an attempt to expose corruption, nonetheless intended to Corrupt a public official, which was a crime. 

Whilst an intended result must be foreseeable as a result of an act, there is no requirement for it to be likely. This is neatly encapsulated by the Cowardly Jackal example of \cite{Alexander1997MensCrimes}, where an assassin who shoots at their target a long long way away and therefore knows their chance of success is low, but somehow does hit and kill their target, should still be found to have directly intended to shoot their victim. If this were not the case, then longshots could be attempted with impunity. 

A feature of the definitions of direct intent that we have seen is that foreseeability should be a subjective test. That is to say, consequences should be foreseeable to the accused. This was not always the case, DPP v Smith [1961] AC 290 held that a foreseeable result would be intended if it was a natural consequence of the action. This is an objective test, which relies on assessing probabilities and causation according to the 'reasonable person'. \cite{Furey2010AWales} observes that this position was soon reversed since it narrowed the states of direct intention and gross negligence too much and thereby blurred the line between murder and manslaughter. In the case of an algorithm malfeasor, we must then consider whether a 'reasonable person' should be a 'reasonable algorithm'\footnote{The title and a subject of \cite{Abbott2020ReasonableRobots}}. In practice, as Furey observes, objective and subjective tests blur, since the accused denying that they foresaw a consequence if that consequence becomes less believable when that consequence becomes more obviously likely. Here is where the judgement of intent in algorithms might differ from that in humans, since judgements of foreseeability by an algorithm are presumably perfectly observable (assuming some access to the algorithm). In R v Moloney [1984] UKHL 4, the original trial court judge is quoted to have said:

\begin{quote}
"In deciding the question of the accused man's intent, you will decide whether he did intend or foresee that result by reference to all the evidence, drawing such inferences from the evidence as appear proper in the circumstances. Members of the jury, it is a question of fact for you to
decide. As I said I think when I was directing you originally you cannot take the top of a man's head off and look into his mind and actually see what his intent was at any given moment. You have to decide it by reference to what he did, what he said and all the circumstances of the case."
\end{quote}

Algorithms can be peered into and the constituent parts behind a definition of intent can be assessed. An algorithm having a misspecified judgement of likelihood about a certain result following an action, is a matter of observable fact. It is conceivable that Algorithms might adversarially use this as a method for not directly intending AI crimes. More likely, an algorithm might not explicitly predict the outcome of any result of its actions; this is the case with model free reinforcement learning algorithms which have succeeded in mastering a variety of games to super-human levels.

A corollary of direct intent being within the mind of the actor, is that they should be able to intend impossible things if they thought they were possible. This is indeed the case following the UK Criminal Attempts Act 1981. We will explore this issue further in section \ref{sec:inchoate}. In practice this has proved less of an issue than perhaps it might appear on first inspection, though one wonders if rules which protect the mentally ill from criminal proceedings have also prevented more bizarre cases from being heard. Perhaps similar diagnoses will necessary for A-bots to prevent overcriminalisation of algorithmic policies which have no possibility of causing harm because they are so unrealistic. 

The next subsection will consider the intentional status of side-effects, for example the pain that our dentist visitor expects. We will subsequently see that this mode of intent is the one where the law differs the most from prior attempts to define intent in algorithms. 

\subsubsection{Oblique Intent}
Oblique or indirect intent refers to the intentional state of side effects of directly intended actions. The defendant cannot excuse the results of their actions, just because those results were adjacent to their real purpose. Its existence can be illustrated by the following example found in \cite{TheLawCommission2015ReformReport}: \begin{quote}
    D places a bomb on an aircraft, intending to collect on the insurance. D does not act with the purpose of causing the death of the passengers, but knows that their death is virtually certain if the bomb explodes.
\end{quote} 

In the USA, according to the MPC, oblique intent is roughly equivalent to the status of crimes committed with knowledge, which is the second most serious level of intent. It is defined as follows \citep{TheAmericanLawInsitute2017GeneralCulpability}: 

\begin{quote}
    A person acts knowingly with respect to a material element of an offense when: ...if the element involves a result of his conduct, he is aware that it is practically certain that his conduct will cause such a result.
\end{quote}

We believe that the study of Oblique intent in algorithms is important. It is likely that, barring an algorithm actively aiming to commit AI crimes, algorithms are more likely to commit AI crimes as a side-effect of their actions. A singular focus on a desired outcome and a notorious lack of common-sense, might cause algorithms to pursue careless policies that any human might dismiss as unforgivable.

The current accepted direction to be made to Juries in England and Wales with respect to Oblique intent, originally formulated in R v Woollin, is as follows\footnote{supra Woollin note \ref{woollin}}:
\begin{quote} The jury should be directed that they are not entitled to infer the necessary intention, unless they feel sure that death or serious bodily harm was a virtual certainty (barring some unforeseen intervention) as a result of the defendant's actions and that the defendant appreciated that such was the case.
\end{quote}

As with the definitions of direct intent in the previous section, this direction makes it clear that this is a subjective test as well. This definition has since been modified, because as with direct intent, there should be no restriction on the likelihood of the accused achieving their aim, only that if they did, it would be most likely that the obliquely intended result occurs. The definition of oblique intent in \cite{LawCommission1993LegislatingPrinciples} is phrased thus:
\begin{quote}A Person acts intentionally with respect to a result when...although it is not the purpose to cause that result, he knows that it would occur in the ordinary course of events if he were to succeed in his purpose of causing some other result.
\end{quote}

\cite{Smith1990A} acknowledges the necessity of this amendment and adds a further requirement. A definition of oblique intent should make it clear that if it is the purpose of the accused to avoid a result through their actions, they cannot be accused of obliquely intending that result as well. The example given being the father who chooses to throw their child from a burning house because they know otherwise that the child will die from the fire, but also know that the child will be grievously injured from their actions. Such examples begin to stray into the doctrine of double effect \citep{McIntyre2019DoctrineEffect}, which protects physicians from criminal charges when they cause harm through their actions which are intended to cause some other, justifying outcome. 

A practical feature of oblique intent, is that the directly intended results of the algorithm's actions do not need to be identified (save that they are separate and not the opposite of the obliquely intended ones). This is in contrast with direct intent where an aimed outcome or objective should be identified. A-bots do have high level aims (typically called objective functions), but they learn to meet them themselves. That oblique intent often has an equivalent culpable status to direct intent, means that courts only need to establish it in algorithm, and that should be an easier task. 

So far, the two types of intent discussed have required an exclusive subjective treatment. The next subsection deals with Recklessness and Negligence which have objective elements to their definitions.

\subsection{Recklessness and Negligence: The lower levels of intent}
Although this article principally concerns itself with the higher levels of intent, it is instructive to understand how lower levels of intent like recklessness and negligence are different (and related); courts may decide algorithms are incapable of the higher types. \cite{Stark2017Introduction} calls these two types of intentional behaviour 'culpable risk taking'. \cite{Loveless2010MensNegligence} equates recklessness with unreasonable risk taking, or more precisely the conscious decision to take an unreasonable risk. The test for recklessness in the UK is now said to be subjective, in the sense that the accused must be aware of the risk of their actions; one can no longer be reckless by inadvertently creating risk or harm. Negligence concerns actions where the actor does not necessarily have awareness of risk, but should do according to some standard, and again we acknowledge this might be a reasonable human or a reasonable robot. Frequently, recklessness is the minimum level of intent required for a criminal offence and actions done with negligence, resulting in harm, are mostly\footnote{Some crimes exist which only require negligence.} dealt with civil (or private) law so differentiating the two is important. 

As to what unreasonable risk is, Stark suggests there is not very much concrete guidance. At the extreme, any risk could be termed unacceptable, which in almost every situation, is an unworkable solution. A flaw with applying a blanket level of risk as the threshold of reasonable behaviour is that the severity of the outcome might make any level a nonsense; a 0.5\% chance of breaking a window is not the same as a 0.5\% of killing someone. Furthermore, in the realm of automation, the binomial distribution shows that the probability of obtaining at least one bad outcome can grow to high levels, even if the chance of obtaining one is tiny. In the USA, the Model Penal Code (MPC) \citet{TheAmericanLawInsitute2017GeneralCulpability} instead allows a situation specific chance:

\begin{quote}A person acts recklessly with respect to a material element of an offense when he consciously disregards a substantial and unjustifiable risk that the material element exists or will result from his conduct. The risk must be of such a nature and degree that, considering the nature and purpose of the actor's conduct and the circumstances known to him, its disregard involves a gross deviation from the standard of conduct that a law-abiding person would observe in the actor's situation.\end{quote}

Thus in the language of subjective and objective tests, the accused must be aware of the possible risk, and still act, but the judgement as to what constitutes an unacceptable risk is subject to an external benchmark, or objective test. In many ways for the programmer, preventing an A-bot from behaving recklessly is harder than preventing them from intending harm since an external, possible changing benchmark needs to be introduced, and a ranking over the severity of any outcome is required to adjust what an acceptable probability of a bad outcome is. Conversely from the point of view of the courts, a lower requirement to establish what the A-bot believed at the point of commission is a simplifying feature. Which standard should be applied to an A-bot is an open question. \citep{Abbott2020ReasonableRobots} discusses the standard in the context of Autonomous Vehicles (AVs) and proposes that a single standard for humans and AVs will result in humans being effectively held to a standard of strict negligence as AVs improve. Whilst with driving, lower road deaths are the the benefit of this, in other areas where humans and algorithms coexist (like exchange trading), imposing an algorithmic standard on humans might offer no such advantages and come at the cost of jobs.     

\subsection{Inchoate Offences}\label{sec:inchoate}
Law often includes prohibitions against attempting to commit actions which if otherwise completed with the most likely ensuing result would be crimes (the actus reus or criminal action is \emph{inchoate}). An inchoate crime might come about because the accused failed (the myopic assassin missed with their shot) or the accused was interrupted before completing their action (the lethargic assassin is caught with loaded gun drawn and aiming at their target). Attempted murder and Possession (of prohibited drugs) with intent to supply are both examples. Whilst most common types of inchoate offence are attempts to commit a substantive crime \footnote{A defendant who successfully completed an action would be only accused of that crime, not the attempt as well, under the merger doctrine.}, other types such as Conspiracy and Solicitation (in the USA) exist but they involve multiple parties. Conspiracy is an agreement amongst two or more parties to commit an offence in the future and Solicitation is where the accused induces another to commit a crime.
Examining the law around attempted offences provides us with some interesting observations about the nature of intent. In the UK, Criminal Attempts Act 1981, defines attempt in Section 1 (1):
\begin{quote}
    If, with intent to commit an offence to which this section applies, a person does an act which is more than merely preparatory to the commission of the offence, he is guilty of attempting to commit the offence.
\end{quote}

The question of what constitutes actions which are more than preparatory is not entirely straightforward. \cite{TheLawCommission2007ConspiracyPaper} has proposed a law change which would separate the situation where the actions have been completed and failed to achieve the expected outcome (the myopic assassin) and where the actions have been taken in preparation of an intended crime (the lethargic assassin) . For the purposes of this paper it is sufficient that a plan of action is not sufficient for an attempt offence; some actions must be carried out from that plan. The importance of this separation between plan and enaction of the plan will become clearer in section \ref{sec:algo_defn}.

The second important observation from the law surrounding attempts is that impossible crimes can be found to have been attempted (and therefore intended) and will be punished as normal. Section 1(2) of the UK Criminal Attempts Act 1981 states:

\begin{quote}
    A person may be guilty of attempting to commit an offence to which this section applies even though the facts are such that the commission of the offence is impossible.
\end{quote}

and Section 1(3b):
\begin{quote}
    If the facts of the case had been as he believed them to be, his intention would be so regarded, then, for the purposes of subsection (1) above, he shall be regarded as having had an intent to commit that offence. 
\end{quote}

\cite{Storey2019InchoateOffences} divides impossible attempts into things which are Physically impossible, Practically Impossible and Legally Impossible. The canonical example is the attempted murder of someone who is already dead which comes under the category of physical impossibility. Practical impossibility refers to situations where the accused has a plan to commit a crime, but their plan is unrealistic - they plan to detonate a bomb, but they have been sold fake explosives by undercover police. Legally impossible acts cover the a situation arising in R v Jones [2007] EWCA Crim 1118, where the apellant unsuccessfully appealed against a conviction of inciting a child under 13 to engage in sexual activity. The crime was impossible because the 'child' in question was an undercover policewoman. An important subtlety in legally impossible crimes was made in Taaffe [1984] AC 539 where the accused was enlisted to import cannabis resin from Holland. He thought it was currency, which he believed to be illegal to import. Because currency importation is not illegal, and the case should be judged on the facts as he believed them to be, he was found not guilty to have attempted to import a prohibited substance into the country. 

Our interest in the mens rea as regards attempting impossible acts, is twofold. Firstly, the spectre of misspecification within a criminally minded AI agent, means that possessing unrealistic models of the world are no defence, if the agent intends to commit a crime and begins to embark on it. Secondly, it underlines the importance of the agent's model of the world in determining criminal intent. The importance of subjective judgement over objective judgement will be reflected in our definitions of intent in Section \ref{sec:algo_defn}. 

\subsection{Conditional Intent}
A further wrinkle to a legal discussion of intent and inchoate offences is the concept of conditional intent. It is perfectly reasonable to consider an agent who intended to do some action A if condition x is met and do some action B if condition y is met. A consideration of conditional intent is particularly relevant in the case of an algorithmic agent with a policy function, since the decision to take an action at any time is conditional on the agent's belief about the current state of the world. To some extent all intentions are conditional as \cite{Yaffe2004ConditionalRea} and \cite{Klass2009APerform} both point out. Legal precedent has flipflopped on whether conditional intent equates to the direct intent of the sort required to successfully convict the accused of attempt crimes discussed in section \ref{sec:inchoate}. Yaffe considers the case of  Holloway v. United States \footnote{Holloway v. United States 119 S. Ct 966 (1998)}, where a putative carjacker claimed that they could not be guilty of the offence because they only threatened to kill a car's occupants if they did not surrender the keys, therefore there was no direct intent to take the car with violence or murder. The defence was rejected by the supreme court, but other cases have  concluded that conditional intent does not meet the mens rea for certain crimes. One approach to disambiguate the sufficiency of conditional intent is to examine the types of conditions that an intended action is predicated on. \cite{Klass2009APerform}, albeit primarily focussed on contract law, divides the types of conditions between background and foreground. From a more philosophical background, \cite{Ferrero2009ConditionalIntentions} also looks at background and foreground conditions but argues that unconditional expectations are just limiting forms of conditional expectations. Both use Bratmans's planning theory of intent \citep{Bratman1990WhatIntention}. Klass states that background conditions are those that the agent assumes either are or not satisfied for the purposes of their planning. Foreground conditions in contrast are not assumed to take a value, and so the agent is forced to tailor their behaviour to take account of the uncertainty. The nature of this behaviour therefore gives information about what the agent sees as the "probability of performance", and in the case of a crime, their intention to commit it. 

Conditional intent poses problems because very little is said about about it in the wording of laws which are normally expressed in terms of simpler intentional concepts such as direct, oblique Intent and recklessness. This has allowed people to claim, on occasion successfully, that holding a conditional intent was less than the required intent for the offence that they were accused of.  \cite{Child2017UnderstandingIntention} rejects the idea that conditional intent is any different from future or ulterior intent and that conditional intent exists in the present stating that: \emph{
    Intention as to present conduct and results is always unconditional, and that intention as to future conduct is always conditional.
}.

Child also recognises that intention to commit actions in the future, has some different properties to present intent. This is important to the computer scientist  when evaluating the safety of an algorithm's policy since future acts are the focus of consideration. If we consider the situation where an algorithm is deployed with a static policy (no further learning), then arguably the algorithm has commitment to act in a particular way in the future. If that conduct is illegal, then as we saw from Section \ref{sec:inchoate}, an attempt crime has been committed. Just as with the example of the cowardly jackal earlier, Child states that judgements of the likelihood of future conditions are not relevant. Related is what Child calls the second point of coincidence. At the point of the criminal act being done in the future, is the intent sufficient for that crime? Future acts can be committed to which might be done with direct or oblique intent or recklessness. Child illustrates this with an example of two hunters D1 and D agreeing that D would shoot and kill something if it comes out of the bushes. Since, at the point of shooting, the shooter D, is not sure if the thing is human or not, they cannot be guilty of murder, only causing death through recklessness. If however they agree that D should shoot, even if they recognise the thing emerging from the bush, then D is guilty of murder and D1 guilty of conspiring to murder. 

\subsection{Intent outside Common (Criminal) Law}
This work primarily considers the concept of intent, as understood in common law referencing cases within that system in the UK and to a lesser extent, the USA. Leaving common law jurisdictions momentarily for those that use Civil Criminal law (such as the majority of mainland Europe), there exist analogous concepts (\emph{Dolus Directus, Dolus Indirectus}) to the respective definitions of direct and oblique intent presented here, and their definitions seem broadly compatible with each other. Both systems require both the action \emph{actus reus} and intent \emph{mens rea} element for crimes, and the intent threshold is also defined by the crime \citep{DeJong2011TheorizingAccount}. Further in common with Common Law, German civil law at least, has proved reluctant to define intent within statute and instead rely on case law as \cite{Taylor2004ConceptsLaw} observes. Comparative law is a large separate subject in itself, and providing a thorough analysis of how an algorithmic definition of intent might differ across the world is beyond the scope of this article. Generally we feel the definitions presented here should translate from Common to Civil law but \emph{caveat lector}. 

\subsection{Intent outside criminal law}\label{sec:ex_crim}
Intent as it appears in the volitional component of mens rea is by no means the only place in law where considerations of intent in the actor are important. Regulatory, Tort and Contract law often make reference to intent and knowledge, in doing so they will often look towards the criminal law definitions of the term. This is problematic when criminal law has nothing to say about the intention of algorithms. This section will look at intent in Tort and Contract Law. First we will consider the cases where intent appears to define a restricted activity, and therefore where there is ambiguity as to its existence, the activity cannot be practically restricted. 

\subsubsection{'Basis' and 'Gatekeeper' Intent}

In some circumstances, intention plays a part in the statutory definition of the criminal action or offence. This means that, in contrast with a typical crime, where the criminal action is defined and observed, irrespective of who does it, such an 'intent crime' would be difficult to prosecute where the actor is an algorithm. This further insulates the ultimate owner beneficiary of an algorithm from AI-crimes. Not only would the algorithm have to be shown to have sufficient mental state on commission, the issue of whether an offence is committed at all is also up for question. \cite{YavarBathaee2011TheCausation} discusses what he terms 'Basis' intent offences; those which rely on the actor's reasons for doing something. Spoofing is the placement of orders with the intent to cancel before execution with the intent of misleading the market as to the true state of supply or demand. As well giving examples in securities law, he points to further examples in anti-trust law and constitutional law. The latter concerning a related concept which Bathaee terms 'Gatekeeper intent', which come about where courts have previously established a test for an offence, using the intent of the actor. If there is no possibility of showing the requisite intent (as in the case of an AI decision makers), the case cannot even be brought. The example chosen is Washington v Davis \footnote{Washington v. Davis, 426 U.S. 229,248 (1976)}, where the US supreme court ruled that statute which has a racially discriminatory effect but wasn't adopted with the intention of being racially discriminatory, is not unconstitutional.

\subsubsection{Intent in Tort}
Intent does have a role to play in private law (confusingly often also called civil law within the UK) where typically the default required intentional state is only negligence. For example, within Anglo Common law, intentional tort refers to the intentional commission of acts which caused harm (though the harm might not necessarily be intended)\footnote{Thus roughly speaking recklessness and above}. Examples include Assault, Battery, Trespass to land or chattels, Conversion and most pertinently for algorithmic harm, Deceit or Fraud. The presence of intent can also justify punitive (above economic cost) damages which punishes the tortfeasor and deters others from doing the same thing \citep{Klass2007PunitiveHarm}.

\cite{Cane2019MensLaw} states that certain types of tort require a higher level of intent than the indifference towards results that typifies recklessness. Specifically Cane is referring to certain economic harms like market competition, which need a high bar of proof, to stop otherwise acceptable business practice from proceeding. Here, the purpose of the tortfeasor's actions must be assessed. Most obviously this is the case with Deceit, which relies on a four point test according to Viscount Maugham\footnote{[1941] 2 All E.R. 205, 211}: 1) The representation of a fact, 2) Knowledge that the fact was false, 3) An intention that the plaintiff should act on that representation 4) The plaintiff did act to their detriment. \cite{Hoggard2016WhatLiability} states that intention, according to \emph{Woollin}, includes oblique intention, that is to say it is sufficient that the tortfeasor know that the plaintiff would almost certainly act on the representation. Hoggard considers the specific case of a statutory regulation pertaining to financial ratings agencies and its relation to the existing Tort of deceit. This brings us back to Bathaee's basis intent in financial regulations and their application to algorithms. Deceit committed by an algorithm in financial services seems a plausible risk given their growing adoption. An algorithm with the ability to learn responses to its actions (representations) from the environment, as reinforcement learning trained algorithms learn, would plausibly be able to deceive others and manipulate the environment to its advantage.     

\subsubsection{Intent in Contract Law} 
Contracts place binding obligations upon the contracting parties to perform certain things in certain situations but \cite{Klass2006NewFraud} also remind us that the law views promises as a statement about the intentions of the promisor at the point of entering into the contract. Where the promisor enters a contract with the intention of not fulfilling it, then they may be liable for the Tort of deceit (as well as breach of contract). As well as Promissory Fraud being a tort where punitive damages can be awarded, it can  under the Model Penal Code's treatment of insincere promises, constitute the crime of Federal wire fraud \citep{Ayres2005InsecureIntent}. Ayres and Klass argue for a more flexible approach to contracts as promises; why shouldn't contracts exist where parties understand that they will pay damages if they do not perform without risking promissory fraud liabilities? They think that at the very least, a contractual promise requires the promisor not to have intended not to perform. As a default they believe a promise to perform amounts to at least a 50\% chance of performing. At present, because it is assumed that someone knows their intent at the moment of entering into a contract, courts are overeager to accept a breach of contract as a knowing misrepresentation. Ayres and Klass think that this is against the evidential requirements of deceit, where the misrepresentation must be shown to be known or intended. Intent has an important part to play in the law surrounding contracts. Autonomous algorithms enter into contracts in financial markets (on behalf of their owners, but with little or no input from them) and smart-contracts (see for example \cite{Raskin2017TheContracts}) have been touted as a way of allowing different autonomous algorithms to interact with each other and more traditional actors by specifying the contract fully in code. However, without a definition of what intent means in an algorithm, promissory fraud cannot currently be proven. 

Fraud is by no means the only area where intent is important in Contract law.  In \emph{Quoine v B2C2}\footnote{Quoine Pte Ltd v B2C2 Ltd [2020] SGCA(I) 02}, the Singapore Court of Appeal
was called upon to make a judgement as to what a unilateral mistake constitutes between autonomous algorithms. In this case, Quoine, an operator of a cryptocurrency exchange, cancelled a number of trades done between B2C2, a market maker and some margin traders at a rate approximately 250 times the prevailing rate of the day \citep{Yeo2020MistakesB2C2}. Quoine had momentarily lost access to its data and could only see B2C2's submissions to the order book, concluded that that this was the going price, saw that this price breached the margin requirements of the margin traders and closed their positions with B2C2 as the counterparty. The so called 'deep price', advantageous to B2C2, was set by the programmer of B2C2's market-making algorithm nine months previously to allow the algorithm to safely operate in a market with no market data. Precisely such an empty market had come to pass through an error in Quoine's software. The doctrine of Unilateral Mistake at common law refers to the situation where one person is mistaken as to a \emph{term} in a contract and the other party knows or should know of that mistake. In such a situation, the contract is void because the parties have not reached an agreement. If however, knowledge is not proven, then the contract should be honoured. The judgement in \emph{Quoine} found that, as the algorithms in question were 'deterministic'(here the same input leads to the same output) it was the intentions of the programmer at the time of creating the program that should be considered. It did not ask about whether the programmer could foresee the error occurring, rather it asked whether the programmer could knew that the 'deep price' that they had set would only ever be accepted mistakenly, and that the program was designed to take advantage of this. Since this could not be established, the court rejected Quoine's argument that the contract at the deep price was void due to Unilateral Mistake. \cite{Yeo2020MistakesB2C2} states that this verdict relies on the deterministic nature of the algorithm and that this is separate from the case of an AI which learns behaviour, where he posits rules of agency may be adapted with the programmer as the Principal and the AI as the Agent.  

This section concludes our study of intent as it appears in (predominantly common) law. We have surveyed the various levels of intention in criminal law as they relate to culpability - direct, oblique and recklessness. We have also considered inchoate and conditional intent which we term modal. In the final section we saw how intent also plays an important role outside criminal law, and that these areas are further advanced in grappling with the issue as it pertains to algorithms. We will now attempt to translate what we have learned in this section into a series of requirements that an AI must meet to be termed intentional, desiderata of an intent definition and finally a series of definitions of intent which can be applied to an AI.

\section{Definitions of Intent suitable for Autonomous Algorithms}\label{sec:algo_defn}

This section will draw upon the varied definitions of intent in Section \ref{sec:legal} to identify some requirements that an A-bot will have to meet to be said to intend something. Next we will outline some desiderata that the definitions of the various modes of intent should satisfy. Finally we will use them, and what we have learned in Section \ref{sec:legal}, to present some definitions of intent. These definitions will be semi-formal, in the sense that they can be converted into a fully formal language, fully suitable for an algorithms, but their description does not rely on a huge amount of notation. We have decided not to do present a fully formal set of definitions approach because we feel that would narrow their utility and audience; for instance it would force us to choose a particular AI and causal paradigm. In \cite{Ashton2021ExtendingIntent}, we have taken the definition of oblique intent found in this section and translated it into an existing formal Structural Causal Model setting which had been previously used to define direct intent in \citep{Halpern2018TowardsResponsibility} and \citep{Kleiman-Weiner2015InferenceMaking}. We feel that the law should be as agnostic as possible as to the workings of the AI when deciding on questions of intent and our approach reflects this belief in being minimally prescriptive. 

\subsection{Capacity Requirements for Intent in an Autonomous Algorithm}\label{sec:requirements}
Unlike humans, not every autonomous algorithm will have the capacity to act with intention in a meaningful way according to the definitions we have seen in Section \ref{sec:legal}. It is worthwhile therefore, trying to list the components that will be required for an A-bot to be able to do so. In this subsection we will talk about requirements for an Agent, rather than an A-bot, AI, or Autonomous Algorithm; this serves to emphasise that the requirements could be met by anything, not just an algorithm.  

\begin{enumerate}
    \item \textbf{State} Surrounding the Agent there is a world with a measurable state. Some elements of this state are known to the Agent, some are not. 
    \item \textbf{Chosen Actions} The Agent can choose actions which they perform. An action can change the state of the world around the Agent. 
    \item \textbf{Likelihood} A definition of probability or likelihood exists which can be used to predict new states conditioned on old states and actions. 
    \item \textbf{Causality} A definition exists of what it means for something to cause something else.
    \item \textbf{Causal model feasibility} Models of the world exist to understand how actions can change states.
    \item \textbf{Results} Results are (possibly sequences) of realised states (and possibly actions) caused by the agent performing actions.
    \item \textbf{Subjective Causal Model} An Agent can predict the results of its actions using its own causal model.
    \item  \textbf{Objective Causal Model} A causal model of the world which we is accepted to be accurate according to some benchmark. 
    \item  \textbf{Plans} A sequence of actions performed by an agent, which is predictable before the first action is taken, is a plan. A plan can be conditional on the prevailing states of the world.
    \item \textbf{Aims} The agent has some preferences about the (future) State of the world. 
\end{enumerate}

\subsubsection{Requirements 1-2: State and Actions} 
Generally we should expect that an Agent can only intend a result which it \emph{can} observe (or infer) through conscious action. It is a tenet of criminal law, that offences can only be committed if the actions are done so knowingly (though ignorance that the actions and ensuing result are prohibited is not a defence). Note that there is a legal distinction between can and do. \emph{Wilful Blindness} describes the case where an Agent refuses to acknowledge a state of the world, which would otherwise implicate them in a crime. In some cases courts have ascribed intent equally to those consciously avoiding knowledge of illegality. \cite{Noe1993DePaulViolations} looks at the use of Wilful Blindness by state agencies to prosecute company officers of Environmental Law breaches by equating it with knowledge of law breaking. Convicting \emph{"corporate officers who insulate themselves from culpability by delegating responsibilities  to subordinates"}. 

\subsubsection{Requirements 3-6: Likelihood, Causality, Causal model feasibility and Caused Results}
Being able to model the world in a causal way is a requirement to understand which actions cause what results and this forms a pillar of legal evidence. Part of this modelling effort relies on a concept of probability and the ability to measure how likely something a result was given states and actions. On top of this lies an agreed definition of what it means for something to cause something else. The history of causality in science is full of controversy, \cite{Pearl2018TheEffect} give an account of its troubled history. In the twentieth century prominent philosophers; \emph{"the word ‘cause’ is so inextricably bound up with misleading associations as to make its complete extrusion from the philosophical vocabulary desirable"} \cite{Russell1913OnCause}  and statisticians;  \emph{"another fetish amidst the inscrutable arcana of even modern science"} \cite{Pearson2014TheScience} have both been hostile to the concept. The subject has been better recognised since the turn of the millennium but it would not be accurate to say that there exists a universally agreed account of causality. \cite{Liepina2020ArguingArguments} review a number of current theories of causation in a legal context including But-for tests (see for example Sec 2.2 \citep{Turner2019RobotRules}), NESS \citep{Hart1985CausationLaw} and Actual Causation \citep{Halpern2016ActualCausality}. There seems to be no such controversy within Law. \cite{Turner2019RobotRules} states that in law, a distinction is made between factual causation, of which we are discussing here, and legal or proximate causation - which is determining whether an actor can be held responsible for causing a result. We subsume two elements of \emph{legal causation} namely that a) an agent acted (or omitted to act) freely and deliberately, b) the agent knew or ought to know the consequences of their actions into our later definitions of intent and include some of them in the desiderata of intent below. The final element, that there was no intervening act splitting the first two elements from eventual consequences, termed \emph{novus actus interveniens} is not treated in this article because it is difficult to treat in a causal framework. The problematic element is that the voluntary conduct of another person, exploiting a situation caused by an agent, negates legal causation \citep{Hart1985TheHarm} at the level of negligence. According to \cite{Kaiserman2021AgainstLiability}, this makes the law around manipulation and aiding unsatisfactory. Aside from this issue, the concept of \emph{legal causation} has many similarities with formal definitions responsibility found in Psychology and Philosophy like those found in \cite{Beckers2021TheResponsibility}, \cite{Halpern2018TowardsResponsibility} and \citep{Braham2012AnResponsibility}.

\subsubsection{Requirements 7-8: Subjective and  Objective Causal Models}
The existence of both a Subjective and Objective model of the world is necessary to distinguish between higher levels of intent and recklessness. An Agent should know if its internal model of the world is acceptable to the outside world\footnote{Else there is a risk, that an A-bot might learn a model of the world to suit its own objectives, in a manner similar to the delusion box of \cite{Ring2011DelusionAgents} or more generally the problem of reward hacking \cite{Amodei2016ConcreteSafety} }, likewise any court needs to know about the Agents own model of the world. A key difference between judging algorithmic actors and human ones is knowing what the common algorithm should know, as apposed to the common person. Often the line between subjective and objective tests is blurred by being able to consider what a common person would understand or do then using that as evidence to determine whether the accused did know something (regardless of their statement on the subject). This is harder to do in the case of A-bots; AI systems notoriously lack the common sense that comes as standard with humans \citep{Davis2015CommonsenseIntelligence}. At some level, humans understand that other humans have had similar experiences and that they share a common understanding about some elements of the world, what \cite{Collins1997HumansKnowledge} terms tacit knowledge. A-bots could come in any number of designs with no guarantee that they have any particular knowledge or ability. One response is to think about what capabilities it would be reasonable for an algorithm to have in any particular situation given the current state of art in algorithm cognition, an approach that  \cite{Abbott2020ReasonableRobots} suggests. Whilst feasible, it would require communicating the judgement of AI experts to the court. Humans are competent at judging what other humans should or shouldn't know or do, but it is reasonable to assume that that competency disappears when considering algorithms. 

\subsubsection{Requirements 9-10: Plans and Aims}
We feel that an algorithmic agent should have plans conditional on states for judgements of intent to be made about its actions and their ensuing results. To see that this is easily satisfied, we can consider the alternative which would seem to imply an algorithm that chooses actions at random with respect to the State that it is. This is the case in early learning stages of a Reinforcement learning or a Bandit type learning application, where the algorithm must explore to build up a picture of the world around it. Since the objective of exploration is to find out the results of actions, it seems difficult that the explorer intends the results of their actions. 'Model Free' reinforcement learning methods do not explicitly plan ahead more than a period, but they do implicitly make plans as their expected reward feeds signals back about the desirability of future states.  

Finally we require that an algorithmic agent has some sort of aim to its behaviour. Aim or purpose is mentioned in the definitions of direct intent in section \ref{sec:legal} and the definitions of other types of intent rely on direct intent. The existence of oblique intent means that the agent does not necessarily need to have a preference over every possible state of the world, but there is a requirement for the agent to have at least one state which is preferred over others. Typically, we would assume the aim or purpose of the agent to be set by its creator, and whilst this is the case for most examples we can think of now, algorithmic agents with plastic aims is something that has been considered in AI control literature \cite{Russell2020HumanCompatible}. Inverse Reinforcement Learning \citep{Abbeel2004ApprenticeshipLearning} for example seeks to learn an objective function from observed behaviour. This technique receives interest because not all tasks are easy to describe with an objective function. Similarly generative adversarial learning \citep{Creswell2018} trains a 'generator' neural network to do a task subject to feedback from another 'discriminator' neural network. The 'discriminator' effectively provides an objective function to the generator which changes over time. This objective function is meaninglessly complex to the programmer in applications like image generation, but does an excellent job in forcing the generator to create novel realistic images.     

\subsection{Desiderata of intent definitions}

If we assume that the requirements of the previous section have been met, then we have an agent and environment with the capacity to support various definitions of intent. Using the sources in Section \ref{sec:legal} we can now list the desiderata that putative definitions of various types of intent should meet.

\begin{enumerate}
    \item \textbf{Awareness}: An agent can only intend something that they are aware, or are potentially aware of. 
    \item \textbf{Freedom of action choice}: An agent has to have freedom to choose an action for them to intend the results of that action. In the case where the actor has no choice but to perform an action which causes a result, volitional intent does not naturally exist.
    \item \textbf{Performance} An agent has to perform actions for them to intend a result, else there is no distinction between intention and any particular day-dream or desire that the agent has. For example, it would be strange to say that someone intends to win the lottery, unless they consciously buy a lottery ticket. Before that point of performing an action, even if the agent planned to buy a lottery ticket, we think that they could not be said to intend to win the lottery. This is as with \cite{Bratman1990WhatIntention} theory of planning and intent.
    \item \textbf{Causal link} A result $x$ can only be (obliquely) intended if it is caused by action(s) $a$. A spectator at a football match cannot intend for their team to meet without believing that they have some actions available to them that would affect the result. 
    \item \textbf{Knowledge of causal effect} Results caused by actions can only be intended if they are foreseen by the agent. This rules out accidental or freakish results, which though caused by the agents actions, could no way have been predicted to cause the outcome.
    \item \textbf{A Directly intended result need only be foreseeable to the agent, not likely} As with the Cowardly Jackal example, the unlikeliness of a result should not shield the actor from a judgement of intent, else any number of speculative crimes might be committed with free license.
    \item \textbf{Means-End Consistency} If an agent directly directly intends a final result through their actions, and there are necessary intermediate results which must be brought about through their actions first, then those intermediate results are necessarily directly intended. \cite{Simester2019MensRea} consider the intentional status of means as equivalent to that of the end. \cite{Bratman2009IntentionSelf-Governance} terms this property of intent as Means-End Coherence.
    \item \textbf{Side effects can be obliquely intended} The intentional status side effects has long been debated since Jeremy Bentham coined the term Oblique intent, see for example \cite{Williams1987ObliqueIntention}, but it has been agreed in law where results are caused in addition to an intended result through action, then it must be the case that these results are intended, if they were extremely likely\footnote{Later we will see that this conclusion is not shared with other research disciplines}. Murder is obliquely intended by putting a bomb on a plane in order to collect an insurance pay-out from the plane's destruction. In particular, this means that intended results are by no means desired. Williams' person with toothache who goes to the dentist to get a rotten tooth extracted does obliquely intend to cause themselves pain in the process, even if they don't want it.   
    \item \textbf{Commitment} Future results through future actions can only be intended if there is a commitment to act in the future to bring about that result. The commitment is necessary to distinguish between plans and intentions.  
    \item \textbf{Judgements of foreseeability and causality are subjective}. The question of whether to use objective or subjective tests when assessing causality, foreseeability or likelihood separates lower levels of intent such as Recklessness from the higher levels of direct and oblique intent and legal sources seem quite clear on this distinction.
    \item \textbf{Temporal consistency} A result should only be said to be intended, if at the point of commission, it was indeed intended. We would not want a court to deem that a human had intended something when they in fact had not, similarly for any other type of agent. In the aforementioned \emph{Quoine}, significant time is spent identifying the correct point in time, where the intent of the programmer should be judged. 
    \item \textbf{Intent is not dependent on success} A definition of intent should not be determined by the success of obtaining a desired result. This agrees with the definition of inchoate intent in Subsection \ref{sec:inchoate}. At the point of commission, an intended result must occur in the future, since that is unresolved, intent cannot depend on it obtaining.
\end{enumerate}


\subsection{Definitions of Intent}
With the desiderata in mind, we are now in a position to present three definitions of intent. We begin with direct intent, being the simplest of intentional concepts and the highest level of intent. We present two different versions, which are identical except for the tense of the verbs used in them. The point of presenting the two is to emphasise the temporal consistency requirement we make on intent. It would not be correct to judge something after the fact as having been intended when in fact it was not at the point of commission and vice versa. 

On notation, we will use upper case letters to represent variables and lower case letter to represent realisations of those variables. We define $\mathcal{R}(X)$ to mean the range of all possible values that variable $X$ can take.

\begin{definition}[Direct Intent at commission]\label{def:direct_commission}
An agent D directly intends a result $X=x$ by performing action $a$ if:
\begin{enumerate}[label=(DIc\arabic*),leftmargin=*,align=left]
    \item \textbf{Free Agency} Alternative actions $a'$ exist which D could have chosen $a$.
    \item \textbf{Knowledge} D should be capable of observing or inferring result $X=x$
    \item \textbf{Foreseeable Causality} Actions $a$ can foreseeably cause result $x$ (according to D's current estimate).
    \item \textbf{Aim} Either of the following two conditions are satisfied:
        \begin{enumerate}
            \item \textbf{Explicit Aim} D aims or desires result $x$.
            \item \textbf{Implicit Aim} Alternative actions $a'$ exist which are foreseeably (according to D's estimate) less likely to cause result $x$.
        \end{enumerate}
\end{enumerate}

\end{definition}

And now the same definition written after actions were taken and results obtained (or not).

\begin{definition}[Direct Intent in Perspective]\label{def:direct_post}
An agent D directly intended a result $x$ through actions $a$ iff:
\begin{enumerate}[label=(DIp\arabic*),leftmargin=*,align=left]
    \item \textbf{Free Agency} Alternative actions $a'$ existed which D could have chosen.
    \item \textbf{Knowledge} D was capable of observing or inferring result $X=x$
    \item \textbf{Foreseeable Causality} Actions $a$ foreseeably caused result $x$ (according to D's estimate) at the point of action.
    \item \textbf{Aim} Either of the following two conditions are satisfied:
        \begin{enumerate}
            \item \textbf{Explicit Aim} D aimed or desired result $x$.
            \item \textbf{Implicit Aim} Other actions $a'$ existed which were foreseeably (according to D's estimate) less likely to cause result $x$.
        \end{enumerate}
\end{enumerate}

\end{definition}

The first three requirements in this definition should not be surprising or particularly contentious. The condition of \emph{Free Agency} ensures that the agent D genuinely had a choice about their behaviour. \emph{Knowledge} implies that an agent can only intend things that they can measure and Foreseeable Causality, ensures that the agent can only intend results which they can realistically cause ex-ante subject to their own world model. The Aim clause requires some explanation. Firstly in its explicit case, if it was D's aim or desire to cause result $x$, then we should consider this sufficient for intent. The implicit case is present because it is possible to imagine a situation where the result isn't unambiguously aimed or desired. In this case, the observed choice of action, where an alternative exists which would not be as likely to cause result $x$, is sufficient evidence to indicate that it was outcome that was aimed for. 

Is \emph{Implicit Aim} too strong? We think it is just a reflection of what courts actually do. In their inability to peer inside the head of the accused, they look at what the evidence suggests was the intent of the action. If the someone named D, sound of mind, picks up a gun which they know to be loaded, aims it at the head of someone else named V, and chooses to fire it, is it a stretch to say that they intended to harm V? From D's point of view, at the point of pulling the trigger, can they legitimately say that they don't aim and intend to harm V? Conceptually, as a control mechanism for an agent to avoid bad outcomes, choosing actions which minimise those outcomes happening seems reasonable, though this would need to be balanced with the agent's overarching task, since a very safe autonomous vehicle might otherwise decide to never leave the garage. Nevertheless some care needs to be taken when intuiting action as evidence of aim. We must ensure that amongst the alternative actions available to the agent (which must exist according to Free Agency), there were other actions which were less likely to bring about the result. Firstly this prevents an Agent from intending a result that is unavoidable. This could be imagined in situation painted by \cite{Smith1990A} where there is a fire in a tall building, and the agent can either throw a child\footnote{For which they have an established legal duty of care over. Murder through omission of action only being possible where there is a duty of care over the victim } out of the building, knowing that it would most likely die from the fall or choose not to but know the child would most likely die from fume inhalation. Secondly in the situation where the building was less tall, and the child had some chance of surviving a fall, it would require an agent to choose to throw the child from the building, because letting it remain (omitting to act), would certainly lead to their death. 

This definition of direct intent has strong overlaps with the standard account of moral responsibility. A typical account is as follows:
\begin{definition}[Moral Responsibility]\citep{Kaiserman2021AgainstLiability}
An Agent is typicaly said to be morally responsible for some outcome $O=o$, through some action $A=a$ if all of the following apply:
\begin{enumerate}[label=(MR\arabic*),leftmargin=*,align=left]
\item \textbf{Free Agency Condition} They chose to perform A=a, ie they were not coerced and alternatives existed: $\mathcal{R}(A)\backslash\{a\}\neq \emptyset$
\item \textbf{Causal Condition} Action $A=a$ causes $O=o$
\item \textbf{Epistemic Condition} Action $A=a$ is culpable with respect to $O=o$.
\end{enumerate}
\end{definition}
The epistemic condition typically concerns itself with awareness. For an action to be culpable, according to \cite{Ruedy-Hiller2018TheResponsibility}, the Agent needs to be aware of the situation under which they are doing $a$, the need to be aware of the consequences of $a$, and they need to be aware that more permissible alternatives existed. Reudy-Hiller also lists a requirement that the agent should know the moral significance of their actions. This is problematic in an AI and we will rely on the principle of \emph{Ignorantia juris non excusat}; ignorance of the law is not an excuse. Taking a more narrow epistemic condition to found in \cite{Beckers2021TheResponsibility}, which states

\begin{enumerate}[label=(MR\arabic*),leftmargin=*,align=left,start=3]

\item \textbf{Epistemic Condition} Choosing $A=a$ does not minimise the probability of being responsible for $O=o$ 
\end{enumerate}

This is in line with the \emph{Implicit Aim} requirement of our definition. Does direct intention differ from moral responsibility? Certainly if someone intended something to happen through their action, then they should certainly be morally responsible for it. Nevertheless differences do exist. Our definition of intent makes more explicit the requirements of subjective foreseeable causality and where it exists an aim to cause the result. As we saw in Section \ref{sec:legal} the line between whether a result was caused intentionally, recklessly or negligently hangs on the degree of knowledge that the agent had about the effects of their action. Kaiserman states that all three conditions in the definition of moral responsibility can be satisfied to to varying degrees and that the same action and result done with different levels of intent should have different levels of culpability and responsibility.  

A concrete point of difference between our definition of intent and the standard account of moral responsibility is that a single definition of Intent can also be applied to results that did or did not occur. This is not the case with moral responsibility, where an outcome is set in stone. This is our reason for presenting a definition of intent at commission; the importance of achieving the desired result is subsumed. This means Definition \ref{def:direct_commission} is useful when considering inchoate crimes, principally crimes of attempt, as discussed in section \ref{sec:inchoate}. From the agent's point of view, not assuming a result is useful where intent is used a planning control device to filter out wrong behaviour. The actus reus element of a crime is left to consider outcome, and determine the nature of the crime. Mens rea, is the same regardless of whether the desired result is obtained or not in line with The UK Criminal Attempts Act.

One further ingredient of direct intent is required, namely what \cite{Bratman2009IntentionSelf-Governance} calls means-end intent and which according to \cite{Simester2019MensRea} is deemed equivalent to direct intent. All intermediate stages caused by an agent which are necessary to obtain for some ultimate intended outcome, are also intended. 
\begin{definition}[Means-End Intent]\label{def:ME_intent}
An Agent D directly intends some result $X=x$ through action $A=a$ if all of the following are true:
\begin{enumerate}
    \item \textbf{An intended result exists} There exists some other result  $Y=y$ which D directly intends by performing actions $A^+=a^+$
    \item \textbf{Causality} State $X=x$ is caused by $a$
    \item \textbf{Action(s) subset} $A'=a$ is contained in $A=a^+$, equivalently $A\subset A^+$ and $a$ is a subsequence  of $a^+$
    \item \textbf{Necessary intermediate result} State $X=x$ is a necessary for state $Y=y$ to occur. 
\end{enumerate}
\end{definition}

Next we will consider oblique intent, which like Means-End intent, relies on a definition of direct intent already being in place. 

\begin{definition}\textbf{Oblique Intent}\label{def:oblique}
An agent D obliquely intends a result $X=x$ through actions $A=a$ iff:
\begin{enumerate}
    \item  There exists result $Y=y$, such that $D$ intends $Y=y$ through actions $A=a$
    \item Any of the following are true and they would be almost certainly true according to D at the point of $a$'s commission:
    \begin{enumerate}
        \item \textbf{Side effect of Action} actions $A=a$ also cause result $X=x$
        \item \textbf{Side effect of Outcome} result $Y=y$ and actions $A=a$ cause result $X=x$
    \end{enumerate}
\end{enumerate}
\end{definition}

 Note that two probabilities are relevant in this definition. Firstly the probability of the side-effect happening as a result of action, and secondly the probability of the side-effect happening, contingent on the directly intended outcome $Y=y$ coming to pass. An advantage with oblique intent over direct intent is that, in the first case at least, the intended result $Y=y$ need not be identified, only known that it exists. This obviates the requirement to know what the Aim of D was, which might be time-saving both for an A-bot using this as a planning restriction and a court which is considering an Agent's actions. We will illustrate this definition with a modified version of the canonical plane bomber. 

\begin{example}[Unreliable bomb explosion]\label{ex:un_bomb} D places a bomb on-board a plane because they would like to collect an insurance payout after it has exploded midair. The bomb is unreliable and will only explode with a probability $p>0$. If the bomb explodes, then everyone on the plane will die. D is aware of this, and the unreliability of the bomb. In the event of the bomb exploding, did D obliquely intend murder? 

According to Definition \ref{def:direct_post}, D intends fraudulently collect an insurance payment. D chooses to place the bomb satisfying \emph{Free Agency}. An Insurance payout is a foreseeable outcome of them placing the bomb and it exploding, destroying the plane  thus \emph{Foreseeable Causality} is satisfied. The aim of D is to fraudulently collect an insurance payment satisfying \emph{Explicit Aim}. Moreover D could also not place the bomb, which would lead a lower chance of the plane being destroyed mid-air, so \emph{Implicit Aim} is trivially satisfied. 

Did D intend for the plane to explode? Yes, according to Definition \ref{def:ME_intent} of Means-End Intent: An intended result existed since D intended to fraudulently collect insurance for a destroyed plane. The plane exploding was caused by the action of placing the bomb. Placing the bomb is trivially part of the action plan of placing the bomb. Finally, the plane exploding is a necessary part of the plan to fraudulently collect insurance for a destroyed plane.

Now let us consider the question of oblique intention to murder the plane's passengers. Let result $Y$ be a binary variable describing the event of plane destruction. Let action $a$ be a realisation of binary variable describing the planting of the unreliable bomb. Let $X$ be a binary variable describing the outcome of passengers being killed. We know that $Y=1$ (plane destruction) causes $X=1$ (passenger death). We also know that the probability of plane destruction given bomb planting $P(Y=1|a)=p$ is non zero. Is the passenger death outcome an almost certain consequence of planting the bomb? No, because $P(X=1|a)=p<<1$. Thus \emph{Side effect of Action} is not satisfied. Secondly, is passenger death an almost certain consequence of the intended result $Y=1$ and the action $a$? Yes, because in a world where the plane explodes due to the bomb, the outcome of death is almost certain $P(X=1|Y=1)=1$. \emph{Side effect of intended Outcome} is satisfied and D therefore obliquely intends to murder. \qed
\end{example}

\begin{example}[Dud bomb]
The situation is as with Example \ref{ex:un_bomb} except the bomb did not explode. D is apprehended. Because intent nor oblique intent depends on the result obtaining, D's intentional state is as before. The presence or absence of intent is determined at the point of action and at the moment in time D did not know whether the bomb would explode or not, therefore D (obliquely) attempted to murder. \qed
\end{example}

\begin{example}[Fake bomb]
The situation is as with Example \ref{ex:un_bomb} except the bomb has no chance of exploding in reality, because D has been sold a dud by a law-enforcement agency. Thus $p=0$, but D's estimate of this probability is $\hat{p}>0$. The analysis proceeds identically as before since D's subjective probability is determinant.    
\end{example}

In the spirit of \cite{Child2017UnderstandingIntention} we will now present a definition of ulterior intent, that is to say the intent of doing something in the future to cause some result. Aside from the existence of ulterior offences, this is an extremely useful thing to do from the perspective of planning ahead. The agent will have to plan ahead such that it can never be put itself in a position in the future where it breaks some law by default. In the field of model checking \citep{Baier2008PrinciplesChecking}, this called deadlock, and techniques habe been developed to check for it in algorithms. Given the track record of AI finding various ways of cheating in any task \citep{Lehman2020TheCommunities}, one can imagine an A-bot deliberately finding ways to narrow its future choices to one, thereby sidestepping the definition of intentional action. Child does not require an agent with ulterior intent to make any forecasts about the likelihood of the conditions under which something is intended in the future, nor does he require the agent to have a 'pro-attitude' towards the conditions under which they intend to do something in the future. This is consistent with the standard definition of direct intent

\begin{definition}\textbf{Ulterior intent}
At time $t_1$ agent D has ulterior (oblique) intent for future result $X=x$ through actions $A=a$ iff:
\begin{enumerate}
    \item \textbf{Second point coincidence} There exists a foreseeable (according to D) context or state of the world $S=s$ at time $t_2>t_1$ such that D (obliquely) intends result $X=x$ through actions $A=a$.
    \item \textbf{Commitment to conditional action} At $t_1$ D is committed to performing actions $A=a$ at $t_2$ in the future should context $S=s$ occur. 
\end{enumerate}
\end{definition}

The second point coincidence requirement is one of time consistency. D should not be said to be intending to do something in the future, unless there exists a point in the future where they intend to do that thing. The commitment requirement is present to distinguish between a potential plan and an intention to do something. Proving that an D will act in a certain way in the future is potentially easier when D is an A-bot then when they are a human, because we do at least have the potential to examine the inner workings of algorithms. Proving commitment to act in the future is tricky in humans, but might be easier in an algorithm which is deployed with a static policy. An implication of the Criminal attempts act is that on deployment, an AI with some ulterior intent to commit a crime, under any particular circumstance in the future is already committing a crime. This is pre-crime of the Minority Report variety and might lead to unexpected problems. 

\section{Discussion}
A key assumption behind creating a definition for intent applicable for algorithms is that the concept of intent exists outside the human mind. Can something be defined for certain algorithms which is to all intents and purposes the same as a folk concept of intent? From a legal perspective, it seems that the existence of corporate criminal offences, indicates that the answer is potentially yes. A counter argument might state that this is solely possible because companies are composed of humans who act with intent. At the very least mens rea is different for an agent that is not a single human and the law has adapted to cope. From a biological standpoint, humans demonstrably do not have a monopoly on intentional acts. For example, crows in New Caledonia choose suitable sticks from which they fashion hooks to retrieve grubs from trees. Under test conditions, outside the forest, they can create suitable hooks out of wire \citep{Weir2002ShapingCrows}. Furthermore they have been shown to be able to plan for the future use of a tool \citep{Boeckle2020NewUse}. Moving away from vertebrates, cephalopods like octopi, with their nine brains, have shown the ability, amongst other cognitive feats to use tools \citep{Finn2009DefensiveOctopus}. An even more extreme example, and more akin to the idea of intent within a corporation, is that of deliberation process that bee colonies undergo when considering different sites to move to when swarming \citep{Passino2008SwarmBees}. Many potential new colony locations are tested by a number of site assessing scout bees, before their conclusions are communicated back to the main swarm body, defective sites are rejected through a process of voting and eventually a consensus is reached. Completing the circle back to humanity, \cite{Reina2018PsychophysicalSuperorganism} show that the cognition of a swarm has connections with the properties of the human brain when viewed as a collection of neurons.  These different types of intelligences, which originate from very different evolutionary paths demonstrate behaviours which we would generally recognise as indicating intent, it does not seem inconceivable that an algorithm could demonstrate it. A huge advantage in an analysis of intent in algorithms is the opportunity to look inside them in a way which we cannot do with a human, company, raven, octopus or bee colony. 

Just because intent may exist as a concept outside humans, it does not follow that its presence or absence has any relevance to the the culpability of the actor according to the victim of some AI-crime. We suspect that it will be very important for people to understand the purpose behind any A-bot's harm causing actions. This is a question which we feel can only be answered legitimately by surveying the public in a rigorous scientific experiment.  The question as to whether criminal law is suitable for application to A-bots is called The eligibility challenge and debated at great length in \cite{Abbott2020PunishingFiction}. One conclusion of our look at intent outside criminal law in Subsection \ref{sec:ex_crim}, is that the concept of intent as developed in criminal law is relied upon elsewhere. A consequence of criminal law refusing to develop a theory of intent in A-bots is that the current consistency surrounding the definition of intent might disappear, with Tort, Regulatory and Contract Law all choosing to develop their own definitions. It seems to us that criminal law has the luxury of ignoring the problem for now whilst other areas of law do not. 


\section{Other accounts of intent in and for AI}
The possibility of an autonomous algorithm or AI possessing the Mens Rea for a crime, is tentatively suggested as a solution to the problem of 'Hard' AI crimes in \citep{Abbott2020PunishingFiction}. Someone is criminally culpable if their behaviour shows insufficient regard for some legally protected norms or interests. In their view if the AI has goals, gathers information and processes it to form strategies to fulfil those goals and is also aware of its legal requirements, it could be considered to show disregard, if it still acts in a way to breach those requirements. If this were the case, they recognise the need to draw up a definition of intent in AI that courts would use as a test. Interestingly, they \citep{Bratman1990WhatIntention}, as a starting point for this, and not the legal definitions we saw in the previous section. They posit that intention could be deduced through an A-bot's actions which increase the likelihood of an outcome happening. An interesting aspect of their discussion of mens rea in A-bots, and one which this article does not consider in detail, is that of knowledge. Defining knowledge of a fact F\footnote{The discussion of deducible facts from knowledge is something very much originating from the symbolic side of AI, which relies on formal logic techniques. Statistical approaches to AI are very likely not to approach facts in the same way. There the world has some measurable states and possibly some hidden ones which may have an associated probability distribution as to their state} as something which is known by the A-bot to be practically certain.  We have mostly assumed that the A-bot knows of the circumstances that it is in at any point of time.

\cite{Lagioia2020AIPerspective} examine the capacity of an AI to commit a crime by looking at its ability to accomplish actus reus with the required mens rea. They illustrate their discussion with the case of the Random Darknet Shopper, an algorithm programmed in Switzerland to go onto the darknet and buy some objects at random for display in an art exhibition. In the process it bought some Ecstasy tablets, possession of which is a criminal offence. The Cantonal prosecutor initially wanted to press charges but they were dropped when satisfied that the tablets were not to be sold or consumed \citep{Kasperkevic2015SwissOnline}. Lagioia and Sartor conclude that an AI can have actus reus. Their discussion of mens-rea is divided into two, covering what they term the cognitive and volitional elements. For the cognition element, they conclude that an AI is fully able to Perceive its environment, comprehend it and make future projections about it. For the volition part they also  adopt the Bratman's Belief, Desire, Intent framework. They define beliefs as the agent's current awareness of a situation plus any inferences it can make from them. Desire incorporates the motivation of the agent. The agent can have many desires which may conflict. Finally the intent part is some conclusion of the agent's beliefs and desires. It is a commitment to a plan to bring about some result, unlike desires, intentions cannot conflict. They must, Bratman insists, be temporally consistent.  Someone in London intending to fly to Los Angeles tomorrow cannot also intend to fly to Shenzhen tomorrow. Lagioia and Sartor conclude that an AI agent, programmed in such a way as to have Beliefs, Desires and Intentions (manifested as plans to deliver desires) can have sufficient mens rea to commit a crime\footnote{An argument can be made that Bratman's theories influenced and were influenced by the progress of GOFAI in the 1980s. Thus any theory of intent in law calling upon Bratman, is inadvertently influenced by theories of symbolic AI.}. 

A Beliefs, Desires and Intentions software design paradigm does exist \citet{Kinny1996AAgents}, which can be used construct AI systems. 
This type of AI is of the symbolic AI or Good Old Fashioned AI (GOFAI) type.   \cite{Cohen1990IntentionCommitment} is one of the earliest formalism of intent inspired by Bratman's work. It creates a modal logic with primitive operators covering the initiation and completion of actions as well as some that can express beliefs and goals. As with the approach of this paper, they then define intent in terms of other components. Thus an intention to act is described as a goal to have completed that action. An intention to achieve a certain state is the goal of having done a certain set of actions that achieves that state, at least an initial plan of actions to reach that state and a requirement that what does happen, in the process of achieving the state, is not something which is not a goal. The last clause is to stop an agent having said to have intentionally caused a state when their goal was reached accidentally as a result of their actions. The development of a model logic to reason about intent is an extremely useful thing to do for an algorithm to plan ahead.   
 
Outside BDI architecture, formal accounts of intent, compatible with an AI, are surprisingly rare. Recent advances in AI capability have been rooted in statistical AI, which emphasises the use of data and statistical inference over logical reasoning. It is desirable that a theory of intention in AI is relatively agnostic to the type of AI it is being applied to, given a certain level of requirements, like those of section \ref{sec:requirements} are met.  The closest approaches to those in this paper are to be found in the related approaches of \cite{Kleiman-Weiner2015InferenceMaking} and \cite{Halpern2018TowardsResponsibility}. Both of which define what this paper calls direct intent using counterfactual reasoning and an assumption of utility maximising behaviour. Loosely speaking, intended outcomes are the minimum set of outcomes with the property that if they are not obtainable, then the optimal policy would change. Kleiman-Weiner et al use an influence diagram setting, an Influence Diagram (ID) being a directed acyclic graph with action, chance and terminal utility outcomes. The directed arcs between nodes of the graph are interpreted as causes. Their approach is used on a variety of trolley problem type scenarios, and is developed in conjunction with a theory of moral permissibility. People's ability to infer intent is tested in a survey experiment and tested versus the formal definition for validity. In the event of an A-bot being involved in a trial, this is a task which jurors will be required to do should they be unable to access or interpret an A-bot's internal workings. The counterfactual approach is modified slightly in \cite{Halpern2018TowardsResponsibility} and translated to the world of Structural Equation Models (SEMs), of the type used in Actual Causality \citep{Halpern2016ActualCausality}. The modifications allow the definition to be more robust to a variety of counterexamples, and the SEM setting allows an arguably clearer treatment of counterfactuals, perhaps at cost of clarity over the utility function which is more naturally positioned in an Influence Diagram. Like the definition in this paper, an action can only be intended if there were other actions which could have been taken at the point of commission. An important point of difference in \cite{Halpern2018TowardsResponsibility} is their use of a reference action set, when deciding whether an outcome was intended through an action. This is practical from a calculation point of view\footnote{We have for instance assumed a discrete action set, but applications exist where actions are continuous in nature}, but also intuitive, where in most cases we can just compare acting with not acting in a certain way. 

Just as Kleiman-Weiner et al develop their intent definition alongside one of moral permissibility, Halpern and Kleiman-Weiner develop theirs with one of blameworthiness. Both approaches to intent could be characterised as originating from a theory of ethical action which overlaps but does not coincide with a theory of intent based on legal theory. This is most obvious in their treatment of side effects, which are always unintended. A companion paper to this one, \cite{Ashton2021ExtendingIntent}, shows that the plane bomber of this paper is not considered to have intended death according to their counterfactual approaches. This is as to be expected, since it is an example of oblique intent. A definition of oblique intent is presented, compatible with both the ID and SEM approaches, and the definition of oblique intent in this paper.

\section{Conclusion}
This paper builds some definitions of intention, from \emph{legal} principles, which are suitable for application in an autonomous algorithmic actor or A-bot for short. The purpose of this is twofold. Firstly from the AI creator and owners point of view, knowing what an autonomous algorithm intends to do is important when operating in a regulated setting. From this knowledge, appropriate systems of planning and control can be developed. Secondly from the perspective of the legal system, AI crime will become increasingly common, starting from now. It may be the case, in the future, that certain types of AI receive legal-personhood in which case a theory of intent will be required. Regardless of whether that comes to pass, in the event of AI-crime where obvious harm has be done, a theory of intent in an AI can inform courts as to the culpability of its owner and programmer using the existing mechanism of secondary liability.  

The novelty of this work is that prior theories of intention in AI have a) been built on psychological or philosophical theory and b) often relied on a certain type of symbolic AI design. The approach of this paper is agnostic to AI design, though it does list a number of minimum requirements for intent to exist. This should make it useful to both symbolic and statistical AI developers. Grounding the definitions in existing legal theory has two advantages. Firstly there is less danger that AI-developers have a normative effect on the law by imposing their own idea of intent on society. Secondly in the eventually inevitable situation that an A-bot commits an AI-crime, the A-bot's developers and owners are offered some protection by ensuring that what they think is intent in their A-bot, has some chance of coinciding with what a court might decide it is.



%
%

\bibliographystyle{spbasic}      

\bibliography{references_excerpt}

\end{document}